# Evaluation of Object Detection Proposals Under Condition Variations

Fahimeh Rezazadegan, Sareh Shirazi, Michael Milford, Ben Upcroft
Australian Centre for Robotic Vision (ACRV)
Queensland University of Technology, Brisbane, Australia
fahimeh.rezazadegan@roboticvision.org

## Abstract

*Object detection is a fundamental task in many computer vision applications, therefore the importance of evaluating the quality of object detection is well acknowledged in this domain. This process gives insight into the capabilities of methods in handling environmental changes. In this paper, a new method for object detection is introduced that combines the Selective Search and EdgeBoxes. We tested these three methods under environmental variations. Our experiments demonstrate the outperformance of the combination method under illumination and view point variations.*

1. Introduction

The issue of object detection started to gain attention in the computer vision community more than two decades ago and recently, there has been a rapid surge towards the improvement of the state-of-the-art detection proposals. Object detection is a fundamental task in many computer vision applications including event analysis, visual tracking, video retrieval and visual surveillance[1]. For instance, object detection is a challenging problem in an automated video surveillance systems, mainly because of the intrinsic object changes (e.g. shape deformation) and extrinsic variations (e.g. occlusion, illumination changes and camera motion). Therefore, it is needed to be done in a reliable and effective way to cope with these challenges with a minimal margin of error on the performance.

Two state-of-the-art object detection approaches, that have been recently attracted a large body of work [2], are Selective Search [3] and EdgeBoxes [4]. Selective Search integrates exhaustive search and segmentation, while the structure of image is exploited to generate the object locations. EdgeBoxes creates a proposal measure by scoring a box based on the number of contours.

In this paper, we introduce a new detection proposal approach by combining these two object detectors. EdgeBoxes is a cost effective detection approach but it is not robust against scene variations whereas Selective Search performs better in expense of higher computational cost. The main incentive of combining these methods is to exploit their advantages and reduce their deficiencies. The new method does not require the highly set parameters of these two approaches which results in a notably lower computational expense, while outperforming those two object detectors. We provide an investigation of Selective Search, EdgeBoxes and the new proposal under severe appearance (illumination) and viewpoint variations as well as varying object sizes.

2. Experiments

This section evaluates the robustness of the mentioned object detectors against the main challenges: Illumination and viewpoint variations and varying object sizes.

2.1. View point variation

In this experiment, we systematically translate a camera in a scene with both close and distant objects once during the day and the second time during the night inducing different lighting conditions. We used the image from the original position as the reference image and move the camera sideways in 22 cm increments. Three images are demonstrated in Fig. 1 as samples.

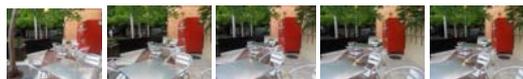

Figure 1: Samples for view point variation experiment.

The employed parameters are σ =1.4, k=1800, nboxes=50 for combination method, nboxes=100 for Edgeboxes and default parameters for Selective Search. In all experiment, the quality of detection is calculated by comparing bounding boxes of detected objects for each method and the ground truth of objects. Comparison is done based on intersection over union (IOU) method [4]. after calculating the quality of detection for several objects, these values are averaged for each image. These average values are plotted with respect to view point variation in unit of centimeters.

The quality of methods which are demonstrated in Fig. 2. reveals that EdgeBoxes is not robust against view point variation. The performance of combination method outperforms other methods by varying camera view point.

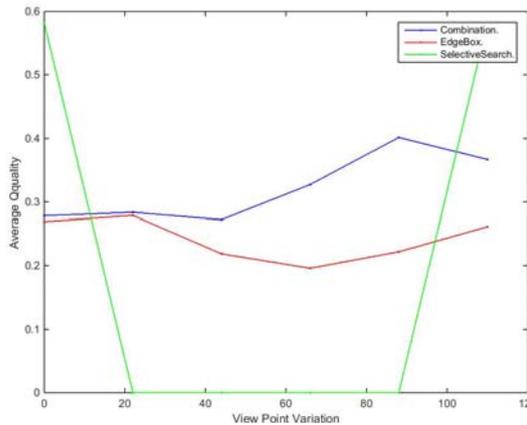

Figure 2: Evaluation of object detection with respect to view point variation

2.2. Illumination and View point

This experiment is conducted based on a dataset including six images taken under the similar conditions of view point variation with illumination difference. Three images are demonstrated in Fig. 3 as samples.

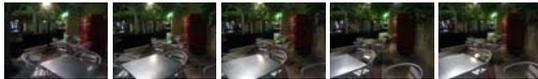

Figure 3: Samples for illumination & view point variation Exp.

Parameters were kept same as previous experiment and same objects were addressed to evaluate the quality of object detection for all methods. The object detection evaluation for three methods are demonstrated in Fig. 4.

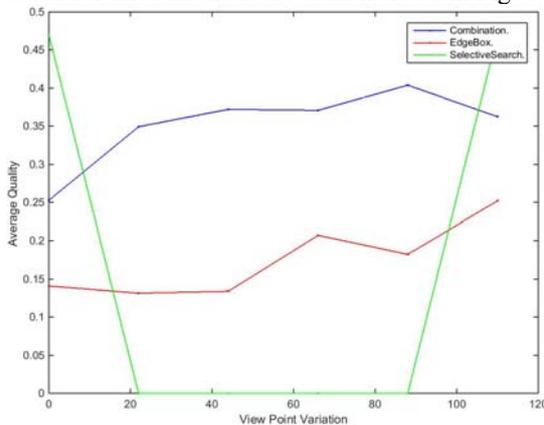

Figure 4: Evaluation of object detection with respect to illumination and view point variation.

The result shows that Selective Search and Combination method works same as previous experiment, but the quality of EdgeBoxes significantly reduced by illumination and view point variation. This will show sensitivity of EdgeBoxes to appearance of edges that are obscured by illumination varying in this experiments.

2.3. Varying Object Sizes

We perform this experiment to quantify the robustness of these approaches against various object sizes. Fig. 5 shows examples of objects evaluated in this experiment. The utilised parameters are σ=1.4 and minSize=k=1800 for Selective Search, nBoxes=100 for EdgeBoxes, and σ=1.4, k=1800, nBoxes=10 for Combination method.

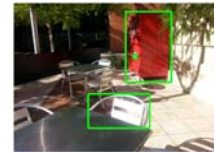

Figure 5: Examples of various-sized objects evaluated.

The quality evaluation for Selective Search, edge box and combination method are demonstrated in fig. 6. The horizontal axis shows objects sizes in descending order.

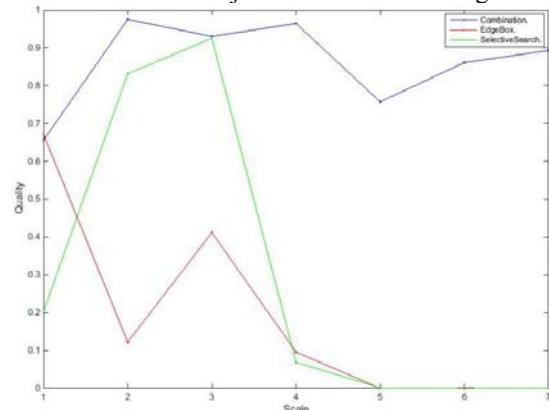

Figure 6: Evaluation of object detection with respect to size.

The result proves that combination of these two methods outperform both of them with same parameters, in particular for tiny objects.

3. Conclusion

Results of experiments shows that new detection proposal can be useful to enhance the robustness of EdgeBoxes against illumination and view point variation and remove the high cost of Selective Search. Further experiments are needed to adjust the parameters of this new method in order to obtain the best performance.